\documentclass{article}

\usepackage[final,nonatbib]{nips_2017}

\usepackage[utf8]{inputenc} 
\usepackage[T1]{fontenc}    
\usepackage{hyperref}       
\usepackage{url}            
\usepackage{booktabs}       
\usepackage{amsfonts}       
\usepackage{nicefrac}       
\usepackage{microtype}      

\usepackage{amsmath}
\usepackage{graphicx}
\usepackage{multirow}

\title{Hyper-dimensional computing for a visual question-answering system that is trainable end-to-end}

\author{
  Guglielmo Montone\\
  Laboratoire Psychologie de la Perception\\
  Universit\'e Paris Descartes\\
  75006 Paris, France\\
  \texttt{montone.guglielmo@gmail.com} \\
  \And
  J.Kevin O'Regan\\
  Laboratoire Psychologie de la Perception\\
  Universit\'e Paris Descartes\\
  75006 Paris, France\\
  \texttt{jkevin.oregan@gmail.com} \\
  \AND
  Alexander V. Terekhov\\
  Laboratoire Psychologie de la Perception\\
  Universit\'e Paris Descartes\\
  75006 Paris, France\\
  \texttt{avterekhov@gmail.com} \\
}

\begin{document}

\maketitle

\begin{abstract}
In this work we propose a system for visual question answering. Our architecture is composed of two parts, the first part creates the logical knowledge base given the image. The second part evaluates questions against the knowledge base. Differently from previous work, the knowledge base is represented using hyper-dimensional computing. This choice has the advantage that all the operations in the system, namely creating the knowledge base and evaluating the questions against it, are differentiable, thereby making the system easily trainable in an end-to-end fashion. 
\end{abstract}

\section{Introduction}
Visual Question Answering (VQA) or Visual Turing Test are terms that refer to a task in which a machine is provided with 
a picture and a question about a picture and the machine is asked to return the answer to the 
question. Such tasks have become popular in the last year thanks to the emergence of 
several datasets containing images, with associated questions and answers \cite{yu2015visual, antol2015vqa}.
Classical deep neural network approaches face these tasks by training architectures composed of 
several parts. In these architectures there are often a RNN (often an LSTM) for encoding the 
question and for producing the answer, and a CNN for analyzing the image \cite{gao2015you, 
ren2015image}. The main idea behind such architectures is to project the question and the image into a 
relatively low dimensional space where the two are compared. These approaches give impressive results when 
the question is about simple properties of the image like "What is the color of the bus?". However the results
become worse when the question is more complex and involves verifying one or more relations among 
objects in the picture ("Is the fruit in the plate different from the one in the basket?").\\
Another group of approaches builds architectures with a \textit{perceiver} and an \textit{evaluator}
\cite{krishnamurthy2013jointly}. The \textit{perceiver} receives the image as input and has to build the knowledge base relative to the input. The \textit{evaluator}, executes the question against the knowledge base to produce an answer. The disadvantage of such 
approaches is that the computations performed by the different parts of the system are of very 
different nature, leading to cumbersome architectures that are complex to train in an end-to-end 
fashion.\\
The main contribution of this work is to show that, at least in the simple case we propose, it is 
possible to build an architecture where the \textit{evaluator} performs computations of the same kind as the 
\textit{perceiver}. It is for this reason that the architecture can be easily trained in an end-to-end fashion. In 
particular in our architecture we will exploit the properties of hyper-dimensional computing (HD-computing) \cite{kanerva2009hyperdimensional}. It is well known in fact that, by defining vectors in a 
high-dimensional space (HD-vectors), it is possible to store and then retrieve information from these
vectors with very simple operations \cite{kanerva2009hyperdimensional}. We will use HD-vectors to store the values of logical constants, and also the values of relations among these constants that will constitute the knowledge base of the system.  In our 
architecture a FFW network will be asked to associate to each image in the input the HD-vector describing the image. The HD-vector returned as output can then be queried to extract answers to given questions, with such questions consisting in sets of differentiable operations on the HD-vectors. The parameters of the network will be updated with a gradient descent procedure in order to minimize the number of wrong answers. In the present paper we prove that this architecture can be successfully trained on a simple VQA task.\\ The paper is organized as follows, in section \ref{sec:dataset} we present the dataset, in \ref{sec:HD} we show how it is possible to encode the knowledge base in an HD-vector and then to retrieve information from it. Finally in section \ref{sec:train} we describe the training procedure and comment on the results of the tests.

\section{Dataset}
\label{sec:dataset}
\begin{figure}[t]
\centering
\includegraphics[scale=0.4]{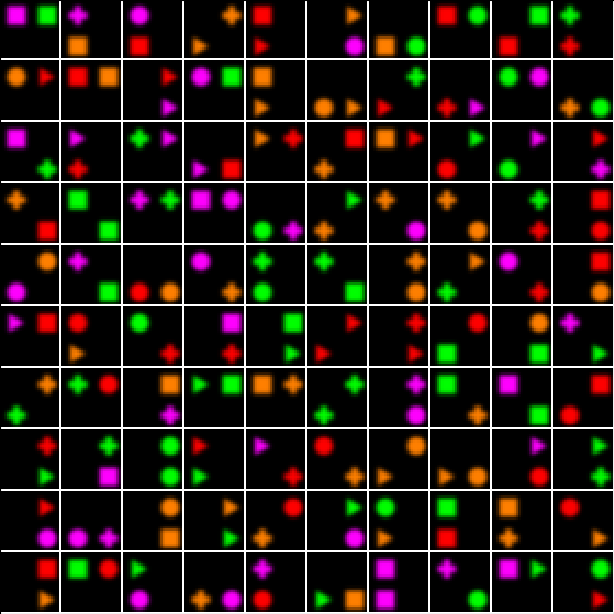}
\caption{Some images in the dataset}
\label{Fig:dataset}
\end{figure}

The dataset is composed of 28x28 RGB images. In each image there are two geometrical figures of four possible colors. The geometrical figures can be in four different positions in the image, namely top-left, top-right, bottom-left, bottom-right. The colors and shapes used are the following:
\begin{itemize}
\item{colors:} red, green, magenta, orange;
\item{shapes:} circle, square, triangle, cross.
\end{itemize}
All possible images containing two figures were created. A sample of the images in the dataset is presented in Figure \ref{Fig:dataset}. The number of total images was 3072. 30\% of these images were used as test set, and the rest of the images were used as training set. 

\section{HD representation of the knowledge base}
\label{sec:HD}
Each picture in the dataset can be described in terms of the following set of natural language concepts: "position", "color", "shape", "top-left",  "top-right",  "bottom-left", "bottom-right", "red", "green", "magenta", "orange", "circle", "square", "triangle", "cross". For example the first picture of Figure \ref{Fig:dataset} can be described by the following sentence: "In position top-left there is a shape of type square and color magenta; and in position top-right there is a shape of type square and color green". In the following we show how by using a method known as HD-computing, it is possible to store and retrieve the information contained in this sentence in one vector \cite{joshi2014language, rahimi2016robust}.\\
We will take the HD-vectors to be 1000 dimensional vectors where each component is randomly chosen to be 1 or -1. We will label HD-vectors using letters in italics, like $k,m,n$. Let's define two operations over the HD-vectors: entangle and grouping. Entangle is defined as follows:
\begin{equation}
\otimes: \quad k = m \otimes n \quad\Leftrightarrow \quad k_i = XOR(m_i, n_i) \quad \forall i \in \{1,\dots,1000\}
\end{equation} 
Please notice that $XOR = (XOR)^{-1}$, so we will sometimes refer to the previous operation also as dis-entangle. Grouping is defined as follows:
\begin{equation}
\oplus: \quad k = m \oplus n \quad\Leftrightarrow \quad k_i = m_i + n_i \quad \forall i \in \{1,\dots,1000\}
\end{equation} 
Also it will be useful to define a distance between two vectors. We choose as distance the following function:
\begin{equation}
cos(m,n) = \frac{\sum_i (m_i \cdot n_i)}{\sqrt{(\sum_i m_i^2)\cdot(\sum_i n_i^2)}} 
\end{equation}
We associate each of the concepts listed in the previous paragraph with a random HD-vector. The HD-vector representing a concept will have the same name as the concept but it will be written in italics. So for example to the concept "color" will correspond the HD-vector \textit{color}.\\ To each image in the dataset we can associate an HD-vector in the following way that we will illustrate with an example. Consider the first picture in Figure \ref{Fig:dataset}. The natural language description of the picture is the following: "In the position top-left there is a shape of type square and color magenta. And in the position top-right there is a shape of type square and color green". To such a description we will associate the following HD-vector:
\begin{equation}
\begin{split}
m  = top\scalebox{0.5}[1.0]{\( - \)} left \otimes (shape \otimes square \hspace{3pt} \oplus \hspace{3pt} color \otimes magenta) \\
\oplus \hspace{3pt} top\scalebox{0.5}[1.0]{\( - \)}right \otimes (shape \otimes square \hspace{3pt} \oplus \hspace{3pt} color \otimes green)
\end{split}
\end{equation}
It can be shown that the information stored in the vector $m$ can be retrieved. For example to retrieve the information about the shape in position "top-left" we can dis-entangle the vector $m$ with the vector $top\scalebox{0.5}[1.0]{\( - \)}left$ and then with the vector $shape$. The resulting vector will be much closer to the vector $square$ than to any of the other three vectors: $triangle$, $circle$ and $cross$.\\ Following the previous example we associated to each image $I^j$ in the dataset an HD-vector $m^j$, building the following dataset:
\begin{equation}
D = \{I^j, m^j\}_{j=1}^{3072}
\end{equation}

\subsection{Querying the knowledge base}
The vectors $m^j$ just defined contain information about the pictures and can be queried to 
retrieve some of the properties of the picture. Querying an HD-vector consists in applying
a set of operations in a specific sequence. In the following we will present the queries we used to train our architecture. The queries will be presented at first in natural language followed by the corresponding set of operations in the space of HD-vectors. To do so let's first define the set \textit{Positions}:
\begin{equation}
Positions = \{ top\scalebox{0.5}[1.0]{\( - \)}left , top\scalebox{0.5}[1.0]{\( - \)}right, bottom\scalebox{0.5}[1.0]{\( - \)}left, bottom\scalebox{0.5}[1.0]{\( - \)}right \}
\end{equation}
In the following the questions.\\
Question 1: "Is there a circle in the picture?"\\
\begin{equation}
\label{eq:question1}
\sum_{pos} cos(circle \hspace{2pt}, \hspace{2pt} shape \otimes pos \otimes m) > 0.5, \quad \quad pos \in Positions.
\end{equation}
Question 2: "Is there the color green?"
\begin{equation}
\label{eq:question2}
\sum_{pos} cos(green \hspace{2pt}, \hspace{2pt} color \otimes pos \otimes m) > 0.5,  \quad \quad pos \in Positions.
\end{equation}

Question 3: "Is there a magenta triangle?"
\begin{equation}
\label{eq:question3}
\sum_{pos} cos(magenta \hspace{2pt}, \hspace{2pt} color \otimes pos \otimes m) \cdot cos(triangle \hspace{2pt}, \hspace{2pt} shape \otimes pos \otimes m) > 0.25, \quad pos \in Positions.
\end{equation}

Question 4: "Is there a square in the bottom-left?":
\begin{equation}
\label{eq:question4}
cos(square \hspace{2pt}, \hspace{2pt} shape \otimes  bottom\scalebox{0.5}[1.0]{\( - \)}left \otimes m)  > 0.5.
\end{equation}

Question 5: "Is the shape in position top-left the same as the one in top-right?":
\begin{equation}
\label{eq:question5}
cos(\hspace{2pt} shape \otimes  top\scalebox{0.5}[1.0]{\( - \)}left \otimes m, \hspace{2pt} shape \otimes  top\scalebox{0.5}[1.0]{\( - \)}right \otimes m)  > 0.5.
\end{equation}

Each of the previous questions has a positive or a negative answer for each of the pictures in the dataset. So to each picture $I^j$ we will associate five numbers, namely $q^j_1$, $q^j_2$, $q^j_3$, $q^j_4$, $q^j_5$. The number $q^j_1$ will be equal to 1 if the answer of Question 1 for the $j\scalebox{0.5}[1.0]{\( - \)}th$ picture is true, otherwise $q_j^1$ will be 0.
We will use these values during the training of the network as target values. So let's add these values to the dataset $D$ previously defined:
\begin{equation}
D = \{I^j, m^j, q^j_1, q^j_2, q^j_3, q^j_4, q^j_5\}_{j=1}^{3072}
\end{equation}

\section{Training and Testing}
\label{sec:train}
We trained a FFW network with two hidden layers of 200 rectified linear units each. The network was asked to return as output the HD-vector describing the picture. For this reason the output of the network was a layer of 1000 nodes with hyperbolic tangent as activation function.\\ The network was trained in order to associate to each input image an HD-vector such that when the vector was queried, it returned the correct information about the image. For this reason the error function that we minimized during the training was composed of a term for each of the questions. In particular for each equation ( \ref{eq:question1}, \ref{eq:question2}, \ref{eq:question3}, \ref{eq:question4}, \ref{eq:question5} ) we took the left side of the equation and computed its value for each image in the dataset by substituting in the equation the term $m$ with the output of the network $net(I^j)$ (here $net$ is the function implemented by the network). The result of the  computation was forced to be closer to 1 if the answer to the question for the picture $I^j$ was true, otherwise the result of the computation was forced to be closer to 0. Let's consider for example Question 1 and its equation (eq. \ref{eq:question1}). Here we defined the error term $E_1$ as:
\begin{equation}
E_1 = \sum_j \left(\sum_{pos} cos\left(circle \hspace{2pt}, \hspace{2pt} shape \otimes pos \otimes net(I^j)\right) - q_1^j\right)^2
\end{equation}
In the same way we defined an error term for each of the questions in the previous paragraph. We called $E_2, E_3, E_4, E_5$ the error terms relative to Question 2, Question 3, Question 4 and Question 5. The error function that we minimized during the training was the following sum:
\begin{equation}
E = \sum_{i=1}^5 E_i 
\end{equation}
We developed two kinds of test for the network. In a first test the network was asked to answer the 
questions used during the training, but on new data. In the second test the network was asked to answer new questions that were not used in the training. In the first test the network was tested on 30\% of the dataset, on data that were not used 
during the training. For each example in the test set we evaluated the answer to each of the questions by 
substituting the output of the net $net(I^j)$ with the vector $m$ in the equation representing the 
question, assuming the answer to be true if the inequality was respected, false otherwise. The 
network reached an accuracy of 100\% on all the questions. On the second experiment we tested the 
network on new questions. In particular we asked questions similar to question \ref{eq:question1}, 
but relative to the three other shapes: "square", "triangle" and "cross". In these cases we obtained  accuracy values of 72\%, 69\% and 60\%.  Interestingly the worst performance was obtained on the "cross" shape, which was the shape that was never used in any of the questions in the training.

\section{Conclusions}
In this paper we presented an architecture for VQA that uses HD-computing to encode the knowledge base and evaluating a query against it. This choice makes the system easy to train in an end-to-end fashion. The system proved to work very well and to generalize well on a simple task. Further experiments are needed to test the architecture on more challenging and natural benchmarks.

\subsubsection*{Acknowledgments}
This work was funded by ERC Advanced Grant Number 323674 “FEEL” and ERC Proof of Concept Grant Number 692765 "FeelSpeech".

\newpage
\bibliography{visualQA}{}

\begin{thebibliography}{1}

\bibitem{antol2015vqa}
Stanislaw Antol, Aishwarya Agrawal, Jiasen Lu, Margaret Mitchell, Dhruv Batra,
  C~Lawrence~Zitnick, and Devi Parikh.
\newblock Vqa: Visual question answering.
\newblock In {\em Proceedings of the IEEE International Conference on Computer
  Vision}, pages 2425--2433, 2015.

\bibitem{gao2015you}
Haoyuan Gao, Junhua Mao, Jie Zhou, Zhiheng Huang, Lei Wang, and Wei Xu.
\newblock Are you talking to a machine? dataset and methods for multilingual
  image question.
\newblock In {\em Advances in Neural Information Processing Systems}, pages
  2296--2304, 2015.

\bibitem{joshi2014language}
Aditya Joshi, Johan Halseth, and Pentti Kanerva.
\newblock Language recognition using random indexing.
\newblock {\em arXiv preprint arXiv:1412.7026}, 2014.

\bibitem{kanerva2009hyperdimensional}
Pentti Kanerva.
\newblock Hyperdimensional computing: An introduction to computing in
  distributed representation with high-dimensional random vectors.
\newblock {\em Cognitive Computation}, 1(2):139--159, 2009.

\bibitem{krishnamurthy2013jointly}
Jayant Krishnamurthy and Thomas Kollar.
\newblock Jointly learning to parse and perceive: Connecting natural language
  to the physical world.
\newblock {\em Transactions of the Association for Computational Linguistics},
  1:193--206, 2013.

\bibitem{rahimi2016robust}
Abbas Rahimi, Pentti Kanerva, and Jan~M. Rabaey.
\newblock A robust and energy-efficient classifier using brain-inspired
  hyperdimensional computing.
\newblock In {\em Proceedings of the 2016 International Symposium on Low Power
  Electronics and Design}. ACM, 2016.

\bibitem{ren2015image}
Mengye Ren, Ryan Kiros, and Richard Zemel.
\newblock Image question answering: A visual semantic embedding model and a new
  dataset.
\newblock {\em Proc. Advances in Neural Inf. Process. Syst}, 1(2):5, 2015.

\bibitem{yu2015visual}
Licheng Yu, Eunbyung Park, Alexander~C Berg, and Tamara~L Berg.
\newblock Visual madlibs: Fill in the blank description generation and question
  answering.
\newblock In {\em Proceedings of the IEEE International Conference on Computer
  Vision}, pages 2461--2469, 2015.

\end{thebibliography}
\bibliographystyle{plain}

\end{document}